\documentclass{trbunofficial}

\usepackage{graphicx}
\usepackage{threeparttable}
\usepackage[utf8]{inputenc} 
\usepackage[colorlinks=true,linkcolor=blue,citecolor=blue,urlcolor=blue]{hyperref}

\usepackage{url}            
\usepackage{booktabs}       
\usepackage{amsfonts}       
\usepackage{nicefrac}       
\usepackage{microtype}      
\usepackage{xcolor}         
\usepackage{subfig}
\usepackage{placeins}
\usepackage{amsmath} 

\AuthorHeaders{Liao, Liu, Kuai, Ma, He, Cao, Stanford, and Ma}
\title{Reconstructing Human Mobility Pattern: A Semi-Supervised Approach for Cross-Dataset Transfer Learning}

\author{%
  \textbf{Xishun Liao, Ph.D.}\\
  Civil and Environmental Engineering Department \\
  University of California, Los Angeles, Los Angeles, California, 90095\\
  xishunliao@ucla.edu\\
  \hfill\break
  \textbf{Yifan Liu}\\
  Civil and Environmental Engineering Department \\
  University of California, Los Angeles, Los Angeles, California, 90095\\
  bmmliu@ucla.edu \\  
  \hfill\break
  \textbf{Chenchen Kuai} \\
  Civil and Environmental Engineering Department \\
  Texas A\&M University, College Station, Texas, 77840 \\
  mobility@tamu.edu\\
  \hfill\break
  \textbf{Haoxuan Ma} \\
  Civil and Environmental Engineering Department \\
  University of California, Los Angeles, Los Angeles, California, 90095\\
  haoxuanma@ucla.edu\\
  \hfill\break
  \textbf{Yueshuai He, Ph.D.}\\
  Civil and Environmental Engineering Department\\
  University of Louisville, Louisville, Kentucky, 40208 \\
  yueshuai.he@louisville.edu\\
  \hfill\break
  \textbf{Shangqing Cao}\\
  Department of Civil and Environmental Engineering \\
  University of California, Berkeley, Berkeley, California 94720 \\
  caoalbert@berkeley.edu\\
  \hfill\break
  \textbf{Chris Stanford, Ph.D.}\\
  Novateur Research Solutions \\
  20110 Ashbrook Place, STE 170, Ashburn, VA 20147 \\
  cstanford@novateur.ai\\
  \hfill\break
  \textbf{Jiaqi Ma, Ph.D., Corresponding Author}\\
  Civil and Environmental Engineering Department \\
  University of California, Los Angeles, Los Angeles, California, 90095\\
  jiaqima@ucla.edu
}

\begin{document}
\maketitle
\section{Abstract}
Understanding human mobility patterns is crucial for urban planning, transportation management, and public health. This study tackles two primary challenges in the field: the reliance on trajectory data, which often fails to capture the semantic interdependencies of activities, and the inherent incompleteness of real-world trajectory data. We have developed a model that reconstructs and learns human mobility patterns by focusing on semantic activity chains. We introduce a semi-supervised iterative transfer learning algorithm to adapt models to diverse geographical contexts and address data scarcity. Our model is validated using comprehensive datasets from the United States, where it effectively reconstructs activity chains and generates high-quality synthetic mobility data, achieving a low Jensen-Shannon Divergence (JSD) value of 0.001, indicating a close similarity between synthetic and real data. Additionally, sparse GPS data from Egypt is used to evaluate the transfer learning algorithm, demonstrating successful adaptation of US mobility patterns to Egyptian contexts, achieving a 64\% of increase in similarity, i.e., a JSD reduction from 0.09 to 0.03. This mobility reconstruction model and the associated transfer learning algorithm show significant potential for global human mobility modeling studies, enabling policymakers and researchers to design more effective and culturally tailored transportation solutions.

\hfill\break%
\noindent\textit{Keywords}: Human Mobility Patterns Modeling, Transfer Learning, Semi-Supervised Learning, Synthetic Mobility Data
\newpage

\section{Introduction}
Understanding human mobility patterns has become increasingly crucial in various fields, including urban planning, transportation management~\cite{wu2019agent,zhang2021effect}, and public health~\cite{pappalardo2016analytical}. As urbanization accelerates and population mobility increases, the ability to accurately comprehend and predict human activity patterns has gained paramount importance. This knowledge not only aids in optimizing urban resource allocation but also provides essential insights for the development of smart cities.

However, current research on human mobility patterns faces two significant challenges. First, most studies rely heavily on trajectory data to analyze spatio-temporal patterns, which often fall short in capturing the underlying semantic interdependency among activities. This approach fails to answer critical questions about human behavior, such as how people schedule their daily activities, what activities typically follow one another, and how activities are distributed within a day (e.g., working and school period). Understanding these semantic relationships is crucial for developing a comprehensive model of human mobility. 

The second challenge stems from the nature of real-world trajectory data, typically collected through GPS-enabled devices like smartphones~\cite{gonzalez2008understanding,laurila2012mobile, pappalardo2013understanding, jurdak2015understanding}. Due to the intermittent nature of data collection and privacy concerns, these datasets often provide incomplete or fragmented views of individuals' daily mobility pattern. This incompleteness makes it difficult to model and understand the full spectrum of human activities and their interdependencies throughout a day or across different contexts.

To address these challenges, innovative approaches that can understand and then reconstruct semantic activity chains are in demand. Such methods must be capable of inferring missing activities, understanding activity dependencies, and capturing the temporal patterns of human behavior. Recent studies have explored annotating human trajectories with semantic information, linking activities to trajectories, which enables the use of natural language processing (NLP) techniques on these annotated trajectories \cite{liu2024semantic,liao2024deep}. Even though with more semantic information, developing these models to learn mobility pattern is still particularly complex when dealing with incomplete datasets, especially in regions where comprehensive ground truth data is unavailable, rendering traditional supervised learning approaches ineffective. In such scenarios, we must explore alternative methods that can leverage incomplete datasets, leading us to consider semi-supervised learning techniques and transfer learning approaches.

In this paper, we propose a model for reconstructing and learning human mobility patterns, focusing specifically on semantic activity chains. This model captures common patterns across agents, learns activity dependencies, and understands the characteristics of each activity, allowing to effectively reconstruct and infer missing parts of activity chains. Furthermore, considering that human mobility patterns vary across different regions due to cultural and environmental factors, and recognizing the challenge of modeling these patterns in areas with limited or fragmented activity data, we propose a novel semi-supervised approach for cross-dataset transfer learning. This learning strategy addresses the limitations of traditional supervised learning, which requires ground truth data, and enables effective cross-dataset and cross-region knowledge transfer. Consequently, this semi-supervised transfer learning approach allows our model to adapt to diverse geographical contexts and overcome data scarcity issues.

With such capability for human mobility reconstruction and knowledge transfer, this model offers significant potential for urban planning, policy development, and transportation system analysis across diverse regions. It serves as a powerful tool for data augmentation, enabling the generation of high-quality synthetic mobility data in areas with limited observational data. This approach facilitates comprehensive data mining and analysis, providing insights into complex mobility patterns. Furthermore, the model's ability to synthesize realistic mobility data significantly advances the field of transportation modeling by enabling the automatic generation of sophisticated simulation models.

Compared to existing literature, our research makes several significant contributions to the field of human mobility pattern analysis:

\begin{itemize}
    \item We propose a generic framework for modeling human mobility patterns across various datasets, regions, and cultures. This framework demonstrates the effectiveness of cross-dataset transfer learning, making it suitable for studying human mobility in data-scarce environments.
    \item We introduce a novel approach to human activity pattern reconstruction, addressing the limitations of current trajectory-based methods.
    \item Our semi-supervised iterative training method enables transfer learning for scenarios lacking ground truth datasets, significantly expanding the applicability of mobility pattern analysis.

\end{itemize}

\section{Related Works}
\subsection{Human Travel Trajectory Reconstruction}
Trajectory reconstruction has become crucial in understanding human mobility patterns, especially when dealing with incomplete datasets. Recent advancements have addressed challenges of data sparsity and irregularity through innovative techniques. Chen et al. \cite{chen2019complete} introduced the Context-enhanced Trajectory Reconstruction (CTR) method, using tensor factorization to reconstruct complete individual trajectories from sparse Call Detail Records. Li et al. \cite{li2019reconstruction} proposed the Multi-criteria Data Partitioning Trajectory Reconstruction (MDP-TR) method for large-scale, low-frequency mobile phone datasets, enhancing reconstruction performance by considering spatiotemporal patterns of missing data and individual similarities. For GPS data, Zheng et al. \cite{zheng2010collaborative} developed a collaborative system for location and activity recommendations, demonstrating significant improvements in inferring activity types. Alexander et al. \cite{alexander2015origin} emphasized the importance of comprehensive temporal and spatial analysis in trajectory reconstruction using mobile phone data.

Despite these advancements, challenges persist, including the diverse characteristics of data sources, limited data accessibility, and restricted model adaptability across different geographic regions. These challenges underscore the need for continued research to develop more robust and widely applicable reconstruction techniques.

\subsection{Transfer Learning }  
Recent advancements in transfer learning have shown significant potential for enhancing human mobility pattern analysis, particularly in scenarios with limited data. Techniques originally developed for natural language processing offer valuable insights for trajectory reconstruction and mobility modeling.

Howard and Ruder's \cite{howard2018universal} gradual unfreezing technique could be adapted to preserve general mobility patterns while adjusting to specific regional characteristics. Peters et al. \cite{peters2019tune} and Merchant et al. \cite{merchant2020happens} suggest that middle layers are often the most transferable, which could be useful when adapting mobility models from data-rich to data-scarce regions. Liu et al. \cite{liu2019roberta} explored freezing bottom layers while fine-tuning top layers, a strategy that could preserve fundamental human movement patterns while adapting to unique characteristics of specific urban environments.

These studies collectively suggest strategies for enhancing trajectory reconstruction and mobility modeling through gradual adaptation of pre-trained models and selective fine-tuning of relevant layers. Future research should focus on adapting these methods to the unique characteristics of mobility data, considering spatiotemporal dependencies and the complex nature of human movement patterns. Such advancements could significantly improve our ability to model and predict human mobility, especially in regions with limited observational data.

\section{Problem Statement}
\subsection{Reconstruction Task}
We denote $i$ for an agent. The $j$-the trajectory collected for the agent contains $N$ stay points,  $Traj_{j}^{i}=\left\{P_{1}^{i,j}, P_{2}^{i,j}, \ldots, P_{N}^{i,j}\right\}$. Trajectories are annotated by one of our previous study \cite{liu2024semantic}, where each stay point is linked with an activity, where each stay point $P_{n}^{i,j}=\left[T_{n}^{i,j},(x,y)^{i,j}, S_{n}^{i,j}, E_{n}^{i,j}\right]$ consists of activity type $T_{n}^{i,j}$ (as shown in Table \ref{table:activity_type_table}), GPS location $(x,y)^{i,j}$, start time $S_{n}^{i,j}$ and end time $E_{n}^{i,j}$. Our goal is to develop a model that can capture the common activity patterns of a region. Therefore, we focus on modeling these activity patterns rather than learning the locations of the agent. An activity chain is a time-ordered sequence of activities, defined based on the annotated trajectory, $A_{j}^{i}=\left\{A_{1}^{i,j}, A_{2}^{i,j}, \ldots, A_{N}^{i,j}\right\}$, where $A_{n}^{i,j}=\left[T_{n}^{i,j}, S_{n}^{i,j}, E_{n}^{i,j}\right]$.

\begin{figure}[ht]
  \centering
  \includegraphics[width=0.98\textwidth]{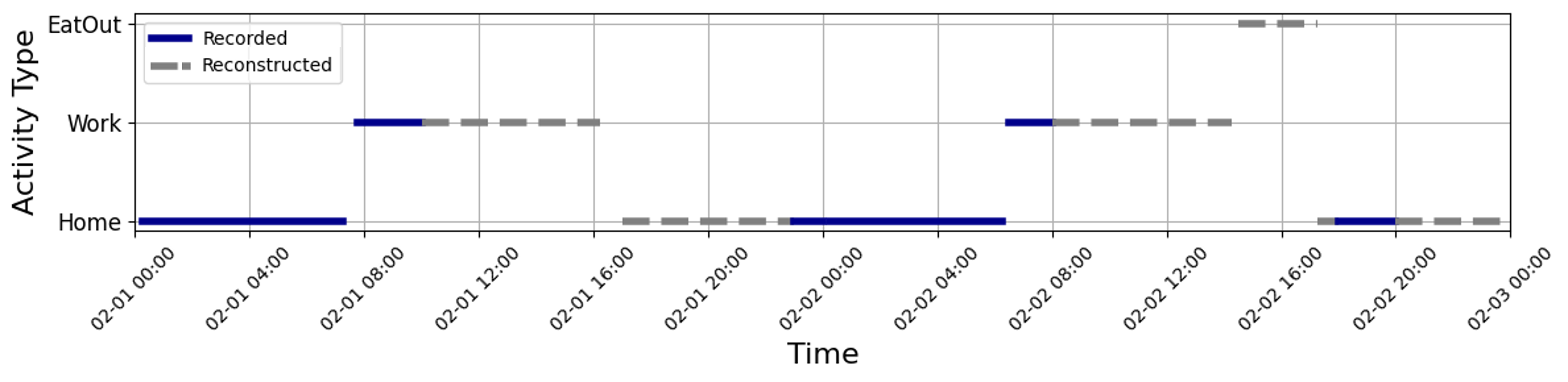}
  \caption{Reconstructed activity chain from an incomplete chain}
  \label{fig:PStatement}
\end{figure}

Due to the fragmented nature of GPS-based data collection methods, stay points usually represent only certain moments of the day rather than covering the entire daily activity of the agent. As a result, the activity chain $A_{j}$ in a real-world dataset is usually incomplete. For example, an observe recorded activities chain (in blue) is shown in the Figure \ref{fig:PStatement}. This chain includes "Home" from "02-01 00:00 to 02-01 07:00", "Work" from "02-01 08:00 to 02-01 10:00", "Home" from "02-02 00:00 to 02-02 04:00", and "Work" from "02-02 06:00 to 02-02 08:00". There are significant gaps in the recorded activities, particularly from "02-01 10:00 to 02-02 00:00" and from "02-02 08:00" onwards.

Finally, provided an incomplete activity chain in region $R1$, a model $M^{R1}$ for region $R1$ can reconstruct possible missing daily activities, as shown by the dashed gray lines in the Figure \ref{fig:PStatement}, filling up the missing time slots based on the common activity pattern of the region. Based on the common activity pattern learned in this region, the model completes the "Work" periods, introduces "EatOut (buy meals)" activities, and proposes "Home" activities to complete the daily cycle, ensuring a reasonable activity chain.
\subsection{Performance Evaluation at System Level}
Given the inherent uncertainty in human behavior, predicting the original activity chain of a specific agent can be challenging and may not always be appropriate. Therefore, our model aims to common pattern, reasonable, align with the distribution of the target region. This approach allows for the reconstruction of plausible full-day activity chains while acknowledging the variability in individual behaviors.

The performance of the model is quantified at the system level by assessing the similarity between the distributions of generated and real-world (ground truth) activity patterns. In this paper, the Jensen-Shannon Divergence (JSD) is adopted as the similarity metric~\cite{luca2021survey}, as presented in Equation \ref{eqJSD}. The modeling objective is to minimize the difference between the distributions of the generated and ground truth activity patterns derived from activity chains. These metrics are 1) \textbf{activity frequencies}, 2) \textbf{start time}, 3) \textbf{end times}, 4) \textbf{number of daily activities} (i.e., length of activity chain), and 5) \textbf{duration of each activity}.

\begin{equation}\label{eqJSD}
JSD(P \| Q)=\frac{1}{2} \sum_{x \in X}\left[P(x) \log \left(\frac{P(x)}{M(x)}\right)\right]+\frac{1}{2} \sum_{x \in X}\left[Q(x) \log \left(\frac{Q(x)}{M(x)}\right)\right]
\end{equation}

\noindent where $M=(P+Q)/2$. Here $P$ represents the distribution of activity patterns extracted from the generated activity chains, while $Q$ represents the distribution of activity patterns computed from ground truth activity chains. $X$ represents the full range of probabilities with respect to a specific activity pattern statistic. As the JSD approaches zero, it indicates a higher level of similarity between the probability distributions being compared, thereby demonstrating the superior performance of model in approximating the true distribution.

\section{Dataset and Data Curation}
\subsection{Dataset with Complete Activity Chain}
The training of the mobility pattern reconstruction model requires the use of complete activity chain, including every activity of an agent in a 24-hour day. Two datasets with completed activity chains were utilized in this study to construct the base model. 

In order to learn common mobility patterns at the national level, we used the 2017 National Household Travel Survey (NHTS), administered by the Federal Highway Administration (FHWA), to capture nationwide mobility patterns \cite{nhtsdata}. The NHTS includes socio-economic demographics and travel diaries from 129,600 households. We aggregated the original 19 activity types recorded in the NHTS into 15 categories based on the activity locations, which are presented in Table \ref{table:activity_type_table}. In this study, only the travel diary of each agent in NHTS is used for training. 

To learn a specific regional activity pattern, we adopted the activity chain data generated by the Activity-Based Model (ABM) of Southern California Association of Governments (SCAG). The SCAG ABM is a comprehensive modeling system that integrates a series of activity-related choice models, ranging from long-term to short-term decisions, to simulate the activity chains of residents in Southern California. This model generates a synthetic dataset that includes detailed information about the types and sequences of activities, locations, and start/end times. These features make the dataset a valuable tool for analyzing the spatiotemporal dynamics of human activities. Further details of the SCAG ABM and its methodology can be found in ~\cite{he2022connected,jiang2022connected}.

\begin{table}[ht]
    \centering
    \caption{Activity category code and their corresponding descriptions}
    \resizebox{\textwidth}{!}{
    \begin{tabular}{|c|p{5cm}|c|p{5cm}|c|p{5cm}|}
        \hline
        1 & Home activities (sleep, chores, etc) or Work from home & 2 & Work-related activity or Volunteer & 3 & Attend school \\ \hline
        4 & Attend child or adult care & 5 & Buy goods (groceries, clothes, gas) & 6 & Buy services (dry cleaners, banking, service a car) \\ \hline
        7 & Buy meals (go out for a meal, food, carry-out) & 8 & General errands (post office, library) & 9 & Recreational activities (visit parks, movies, bars) \\ \hline
        10 & Exercise (jog/walk, walk the dog, gym, etc) & 11 & Visit friends or relatives & 12 & Health care visit (medical, dental, therapy) \\ \hline
        13 & Religious or community activities & 14 & Something else & 15 & Drop off/pick up someone \\ \hline
    \end{tabular}}
    \label{table:activity_type_table}
\end{table}

\subsection{Real-World Trajectory Dataset Processing}

\subsubsection{Raw Data to Trajectory}

\begin{figure}
  \centering
  \includegraphics[width=0.65\textwidth]{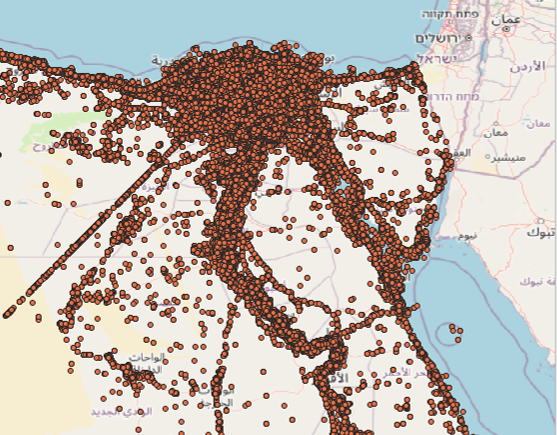}
  \caption{Human travel stay points in Egypt}
  \label{fig:staypoints}
\end{figure}

In this study, we utilized GPS data collected in Egypt, provided by Veraset \cite{veraset}. This data, sourced from devices such as smartphones and tablets equipped with location-based applications, exhibits inherent randomness due to factors such as GPS drift and measurement error. To address this, we developed a methodology to identify stay points to form the trajectory data of each agent, as shown in Figure \ref{fig:staypoints}.

Each record in the raw data is a tuple containing the user identification number, time stamp, longitude, and latitude. We adopt a 2-step process to extract stay points from the raw data. Because a record can be indicative of either a stay or movement, we first label records that are either 5 minutes apart in time or 300 meters away in distance from the previous record as stays. Additionally, we calculate the speed of the movement from the difference in both space and time between consecutive records. Records that show a speed less than 30 kilometer per hour are kept as stay points. 

The second step is to cluster all remaining records, which are now labeled as stay points, into stay regions. The aggregation combines nearby stay points that are presumably a user's visits to the same location. We adopt the clustering method outlined in \cite{zheng_collaborative_2010}, in which the entire region covered by the dataset is divided into cells in a grid system, and the cells in the grid are assigned to various stay regions according to the stay points detected. For this study, the resolution of the stay regions we use is between 0.06378 $km^{2}$ and 0.12709 $km^{2}$ (level 9 hexagons in H3). 

\subsubsection{Activity Chain Construction by Trajectory Annotation}
Based on the Point-of-interest (POI) data provided by Open Street Map \cite{OSM} and GPS trajectories identified as stay points, we create the activity chain by annotating each stay point with a corresponding POI and activity. The process begins by using an LLM to match POIs with probable activities; for example, a restaurant might be associated with activities like 'buy meal' and 'work'. These categorized POIs with activities are then matched to each GPS-identified stay point, taking into account the characteristics of POIs within a 25-meter radius and activities~\cite{liu2024semantic}. 

This whole activity chain construction process comprises two steps: first, the inference of mandatory activities such as Home, Work, and School; and second, the inference of non-mandatory activities. Since mandatory activities typically exhibit relatively consistent locations and periodic patterns, we employ a rule-based method to infer them, as outlined in Alexander~\cite{alexander2015rule}. For instance, the 'Home' activity is identified by selecting the stay point that has the highest visit frequency during night hours across up to six months. On the other hand, we employ a Bayesian-based algorithm to identify non-mandatory activities at each stay point. 

This algorithm ~\cite{liu2024semantic} evaluates the probability of each non-mandatory activity type $T_{i}$ linked to POIs at a stay point, by considering factors such as the activity start time and the characteristics associated with involved POIs. To construct the activity chain, we assign each stay point with the activity with highest probability, as the equation below:

\begin{equation}
T = \arg \max_{T_i} \left\{ P(T_i \mid \text{POI}, S)\right\} = \arg \max_{T_i} \left\{\sum_J P(T_i \mid \text{POI}_j, S) \cdot P(\text{POI}_j) \right\}
\end{equation}
where $T$ is the activity at the stay point, given $J$ POIs and start time $S$.

\subsection{Activity Chain Encoding and Masking}

\begin{figure}
  \centering
  \includegraphics[width=0.9\textwidth]{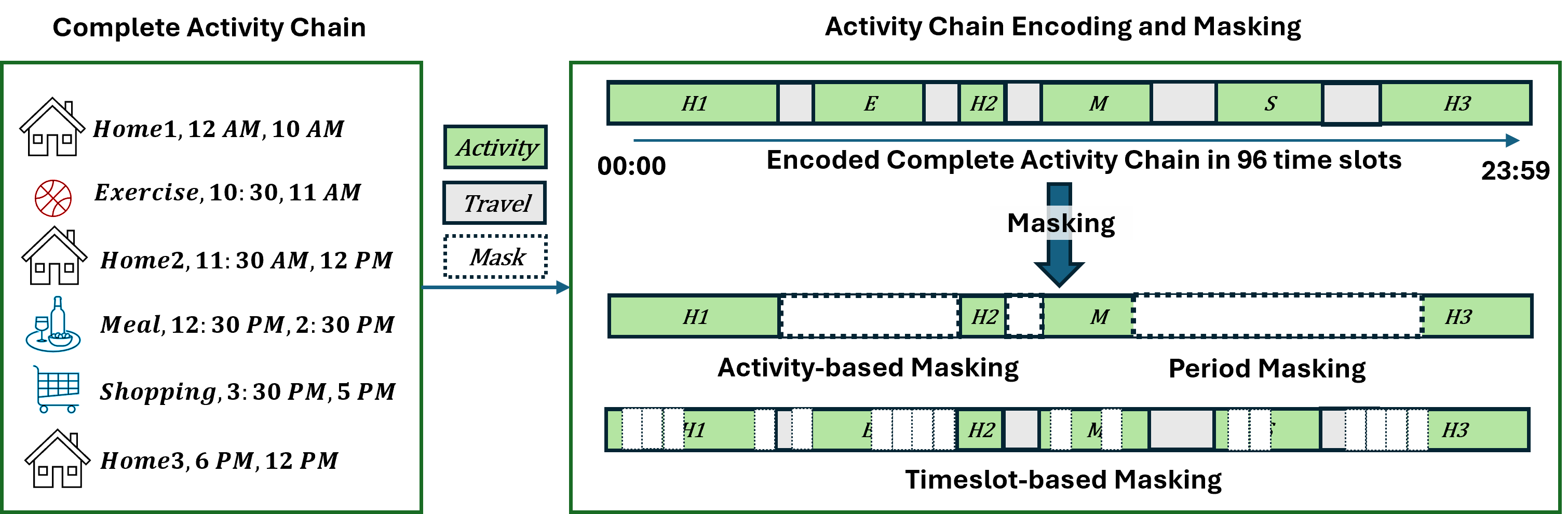}
  \caption{Activity encoding and masking approaches}
  \label{fig:EncodingMasking}
\end{figure}

To effectively process and learn from activity chains, we implement an encoding and masking strategy that allows our model to handle diverse activity sequences and simulate incomplete data scenarios. We encode a day's activity chain into a sequence of 96 time slots, each representing a 15-minute interval. This fine-grained representation allows us to capture detailed temporal patterns of activities. Each activity in the chain is assigned to its corresponding time slots based on its start and end times. For instance, in the Figure \ref{fig:EncodingMasking} shown, "Home1" occupies the slots from 00:00 to 10:00, followed by a 30 minute travel time, then "Exercise" from 10:30 to 11:00, and so forth.

To train our model to reconstruct missing parts of activity sequences and enhance its robustness to incomplete data, we employ three distinct masking strategies that simulate common data missing scenarios encountered in real-world datasets. The first strategy, \textbf{activity-based masking},  involves randomly masking entire activities within the chain, simulating scenarios where specific activities are entirely missing from the data. The strategy, \textbf{period masking}, masks continuous periods of the day, irrespective of activity boundaries, mimicking situations where data for extended periods is unavailable. The third approach is \textbf{time slot-based masking}, which randomly masks individual time slots throughout the day, representing sporadic data loss or intermittent data collection.

These masking strategies simulate incomplete real-world data, enrich training examples, and push the model to robustly represent activity patterns and their interdependencies. By encoding and masking complete activity chain datasets, we generate a diverse corpus that enhances the model's ability to reconstruct activities comprehensively. This method is important for handling incomplete data and transferring knowledge to data-sparse regions. Additionally, applying various masking techniques to the same activity chain substantially increases the training data volume, improving the model's predictive performance and generalizability across different patterns and contexts.

\section{Methodology}

\subsection{System Workflow}

\begin{figure}
  \centering
  \includegraphics[width=0.9\textwidth]{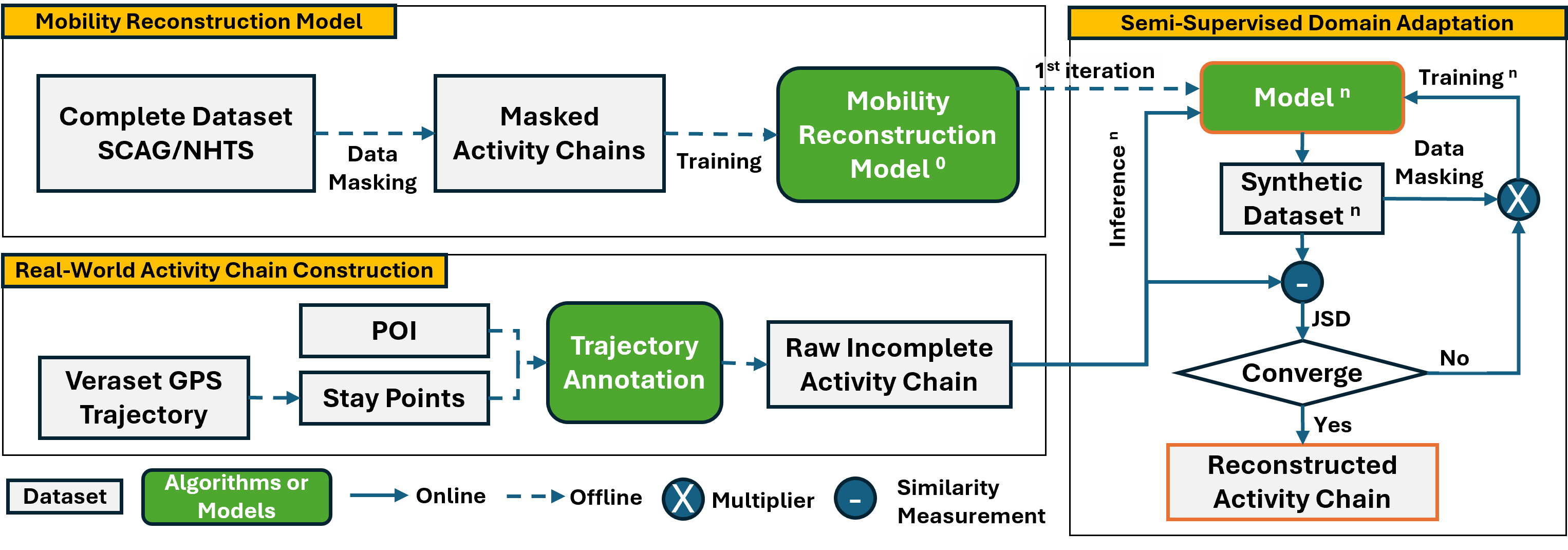}
  \caption{System Workflow for building transferable mobility reconstruction models}
  \label{fig:SystemWorkFlow}
\end{figure}

The workflow of the proposed mobility reconstruction system is structured to optimally leverage comprehensive household travel survey data for application in less data-rich environments, as illustrated in Figure \ref{fig:SystemWorkFlow}. It integrates the principles of semi-supervised domain adaptation in a novel methodological framework that does not rely on ground truth data.

The process initiates with the \textbf{training of the Mobility Reconstruction Model} $M^{0}$ using a comprehensive dataset (e.g., NHTS and SCAG dataset) from data-rich regions like the United States, incorporating detailed activity chains from household travel surveys. This model undergoes training with various masking techniques that mimic scenarios of data incompleteness, enhancing its ability to handle real-world data variations effectively. After training, \textbf{real-world activity chains construction} occurs through data mining techniques applied to GPS trajectory data, focusing on extracting stay points and annotating trajectories with contextual data from points of interest (POIs). These chains remain inherently incomplete, mirroring typical data gaps in datasets. The \textbf{semi-supervised domain adaptation} module then starts, introducing a fine-tuning strategy to adapt the base mobility reconstruction model to regions with limited and incomplete data, bypassing the need for ground truth data. 

This process, which begins by leveraging a complete dataset from a data-rich region and then transferring this knowledge to regions with limited or fragmented activity data, embodies an innovative iterative process of data synthesis and model refinement. It not only facilitates the robust prediction of activity patterns in regions lacking comprehensive data but also enhances the model's generalizability across various geographical contexts.

\subsection{Model Design}
\subsubsection{Model Structure}
\begin{figure}[ht]
  \centering
  \includegraphics[width=0.4\textwidth]{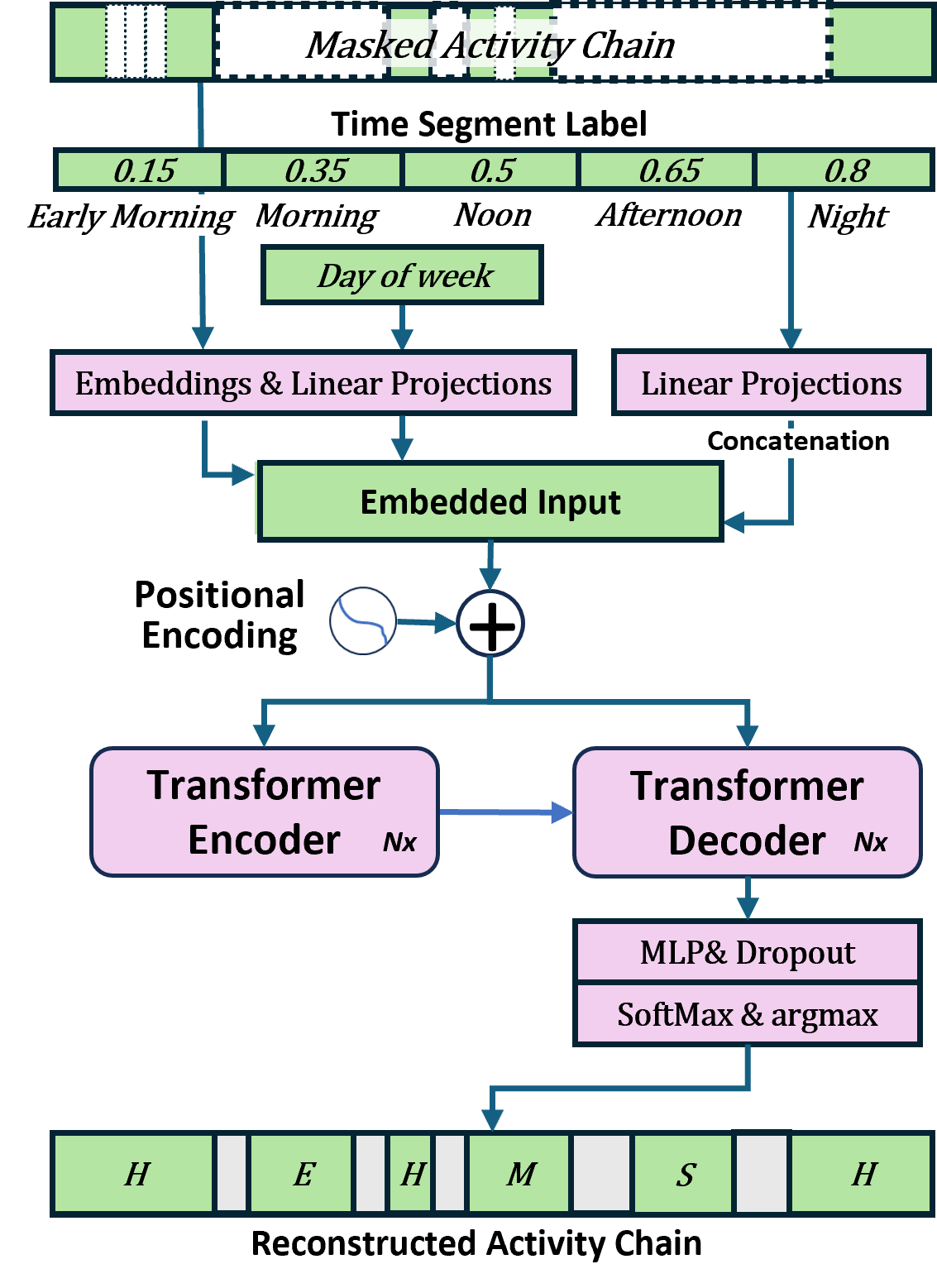}
  \caption{Network architecture of the mobility reconstruction model}
  \label{fig:ModelStructure}
\end{figure}

The core of our approach is a sophisticated model architecture designed to reconstruct activity chains from masked inputs, leveraging the power of Transformer-based structures \cite{vaswani2017attention}. This architecture, illustrated in Figure \ref{fig:ModelStructure}, is particularly well-suited for sequence-to-sequence tasks and incorporates several key components to enhance its effectiveness in handling temporal and contextual information inherent in human mobility patterns.

\textbf{Input Preparation and Enrichment}. The model processes masked activity chains that simulate real-world data incompleteness, enhanced with temporal context through a detailed labeling scheme. Time segments are numerically valued to represent different times of day, from early morning (0.15) to night (0.8), illustrating diurnal patterns. Additionally, day-of-the-week labels capture weekly cycles, essential for analyzing human activity rhythms.

\textbf{Embedding and Feature Concatenation}. 
To optimize processing in the Transformer framework, we utilize an advanced embedding strategy. Categorical inputs, such as activity types and day-of-week indicators, are converted into dense vectors via embedding layers. These embeddings, along with time segment labels, are projected linearly to match the Transformer’s dimensional requirements. The resulting vectors are then concatenated, forming an input vector that captures both temporal and contextual dimensions of the activity chain, setting a robust basis for further processing.

\textbf{Transformer-based Sequence Processing}. The model features three stacked Transformer blocks, each comprising an encoder and a decoder. The encoder uses self-attention to analyze embedded inputs, capturing complex dependencies across the activity chain segments beyond the limits of recurrent neural networks. The decoder then uses this processed contextual and temporal information to predict activity sequences, essential for accurately reconstructing the activity chain.

\textbf{Output Refinement and Activity Chain Reconstruction}.
The decoder's output is refined by a multi-layer perceptron (MLP) with integrated dropout layers, which fine-tunes predictions and reduces overfitting, improving model generalization. A SoftMax layer then converts the MLP's logits into a normalized probability distribution, representing potential activity types for each time slot.

The culmination of this process is a fully reconstructed activity chain, with predictions for both masked and unmasked time slots. This reconstructed sequence not only serves as a direct comparison to the original input chain but also demonstrates the model's capacity to infer missing data and capture the underlying patterns of human mobility.

\subsubsection{Loss Function Design}
Given the encoding format of 96 time slots to describe daily activities, reconstructing the incomplete activity chain can be seen as a classification task for each time slot. The loss function tailored for the task of activity chain reconstruction. This loss function incorporates several elements to capture various aspects of the reconstruction problem, ensuring that the model learns not only to predict activities accurately but also to capture temporal dependencies and transition patterns inherent in human mobility data. The loss function comprises the following components:

\textbf{Cross-Entropy Loss} is commonly used to measure the dissimilarity  between predicted probability distributions and true activity labels to minimize activity mismatch in each time slot.
 
\begin{equation}
L_{CE} = - \sum_{c} w_c \cdot y_c \log(\hat{y}_c)
\end{equation}
where $y_c$ are the true labels for class $c$, and $\hat{y}_c$ are predicted probabilities of for class $c$. The use of class weights $w_c$ allows for balancing the loss across potentially imbalanced activity classes.

\textbf{Transition Loss} focuses on accurately predicting activity transitions by comparing predicted changes in activities with the true changes using binary cross-entropy. It is designed to capture temporal dynamics by encouraging predicting changes in activities at the correct time points by penalizing incorrect transitions. This helps generate more realistic and coherent activity sequences, rather than frequently jumping between activities without any pattern.
\begin{equation}
L_{TR} = -\frac{1}{N} \sum_{i=1}^{N} [t_i \log(\hat{t}_i) + (1-t_i) \log(1-\hat{t}_i)]
\end{equation}
where N is the number of time steps, $t_i$ and $\hat{t}_i$ are true and predicted transitions at time step $i$ (1 if activity changed, 0 otherwise).

\textbf{Dynamic Time Warping Loss (DTW)}, $L_{DTW} = DTW(\hat{y}, y)$, is adopted to evaluate similarity between predicted sequence $\hat{y}$ and true sequence $y$ using DTW distance \cite{muller2007dynamic}.

Finally, the loss functions $L$ combines three loss terms as below:
\begin{equation}
L=w_1 \cdot L_{CE}+w_2 \cdot L_{TR}+w_3 \cdot L_{DTW}
\end{equation}

\subsection{Semi-Supervised Iterative Transfer Learning}
The proposed transfer learning approach is underpinned by research indicating that over 90\% of human activity patterns share similarities across different regions, captured by a limited set of human motifs~\cite{schneider2013unravelling,cao2019characterizing}. This fundamental similarity in human behavior patterns provides a solid foundation for transferring knowledge from data-rich regions to those with limited or incomplete data. The challenge lies in effectively adapting the model to target regions while preserving the universal patterns of human behavior gleaned from the source domain. 

As presented in Figure \ref{fig:SystemWorkFlow}, the base model $M^{0}$ synthesizes the first dataset $SD^{0}$, initiating an iterative cycle of model refinement. Each iteration involves the generation of a synthetic dataset $SD^{n}$ from the raw incomplete activity chains, which serves to train subsequent models $M^{n+1}$. This cycle continues until the JSD between the synthetic and real-world data converges or meets a predefined standard, signaling the adequacy of the model to reconstruct activity chains accurately and its readiness for application in new regions.

\subsubsection{Progressive Unfreezing Approach}
Our fine-tuning strategy employs a progressive unfreezing approach, balancing the preservation of fundamental patterns from the source domain with adaptation to the target domain. Based on the model structure described in Figure \ref{fig:ModelStructure}, comprising embedding layers, Transformer, and MLP, we implement the following phased unfreezing:

In the \textbf{Initial Phase} (first quarter of epochs), only the MLP and the embedding layer are unfrozen. The MLP, being closest to the output, is responsible for the final activity classifications. Unfreezing this layer allows the model to quickly adapt its decision-making process to the new domain. Simultaneously, unfreezing the embedding layer enables fine-tuning of the basic semantic representations of activities and temporal information, such as new features, new activity definitions, and temporal patterns. 

During the \textbf{Intermediate Phase} (second quarter of epochs), we additionally unfreeze the layer of the Transformer's encoder and decoder closest to the input. These layers, being nearest to the input, is generally responsible for processing the most basic patterns in the sequence data. By unfreezing it last, we ensure that the fundamental knowledge learned from the source domain is preserved as much as possible while still allowing for subtle adjustments to better fit the target domain's data characteristics.

In the \textbf{Final Phase} (remaining epochs), we lastly unfreeze the middle layer of the Transformer's encoder and decoder, which typically capture intermediate-level patterns in the data. By unfreezing it, we allow the model to adjust its representation of more complex inter-activity relationships and temporal dependencies that may be specific to the target domain.

This strategy is grounded in the principle that layers closer to the input learn more general, transferable features, while those closer to the output learn task-specific features \cite{yosinski2014transferable}. By keeping input-near layers of the Transformer frozen, we preserve the universal understanding of human activity patterns. The middle layers, typically most transferable \cite{peters2019tune}, are unfrozen last to balance adaptation and knowledge preservation. Gradual unfreezing allows the model to capture increasingly subtle differences between domains while mitigating the risk of catastrophic forgetting. This approach aims to achieve an optimal balance between leveraging knowledge from the data-rich source domain and adapting to the specific characteristics of the target domain.

\subsubsection{Training Strategy}
In addition to the progressive unfreezing approach, we employ several techniques to maintain a stable learning process and enhance the model's performance.

First, to avoid overfitting to the new synthetic data and ensure robustness, we retain 20\% of the previous iteration's dataset in each new training cycle. This retention strategy allows the model to continuously refine its understanding by integrating information from both new synthetic data and a portion of the previously learned data, promoting stability and preventing drastic shifts in learned patterns.

Additionally, real and synthetic data is differentiated during the training process. Specifically, we apply a mask-based loss weighting technique to value the contribution of real data more than synthetic data in the loss computation based on the cross-entropy loss term  $L_{CE}$, as follows:
\begin{equation}
\begin{aligned}
&L_{real} = \frac{1}{N_r} \sum_{i=1}^{N} m_i \cdot L_{CE_i}, \\
&L_{synthetic} = \frac{1}{N_s} \sum_{i=1}^{N} (1-m_i) \cdot L_{CE_i}    
\end{aligned}
\end{equation}
where $m_i$ is the mask (1 for real, 0 for synthetic), $N_r$ and $N_s$ are the numbers of real and synthetic points. Then the total loss function becomes:

\begin{equation}
L= w_1 \cdot (w_l \cdot L_{real} + w_s \cdot L_{synthetic})+w_2 \cdot L_{TR} + w_3 \cdot L_{DTW}
\end{equation}

\section{Experiment and Results}

\subsection{Mobility Pattern Reconstruction}
\subsubsection{Training Details}
Training the base mobility pattern reconstruction model requires complete dataset, such as SCAG ABM dataset and NHTS data. In this study, SCAG ABM data is selected for training the base model, due to its substantial dataset size, in this study we used 700,000 samples for training, 200,000 samples as validation, and 100,000 samples for testing. The choice is necessitated by the data-intensive nature of transformer models. In contrast, the NHTS dataset, containing only 180,000 samples, proves insufficient for effective training based on our preliminary experiments \cite{liao2024deep}.

The training protocol encompassed 120 epochs with a batch size of 512. To mitigate potential overfitting, we implemented regularization techniques, including dropout, L2 regularization, and early stopping. Specifically, the base model training process is stratified into three distinct phases, i.e., 1) Model warm-up using unmasked data for 5 epochs. 2) Training on 40\% masked data for 40 epochs. 3) Training on 70\% masked activity chain for the remaining epochs.

\begin{figure}
  \centering
  \includegraphics[width=0.98\textwidth]{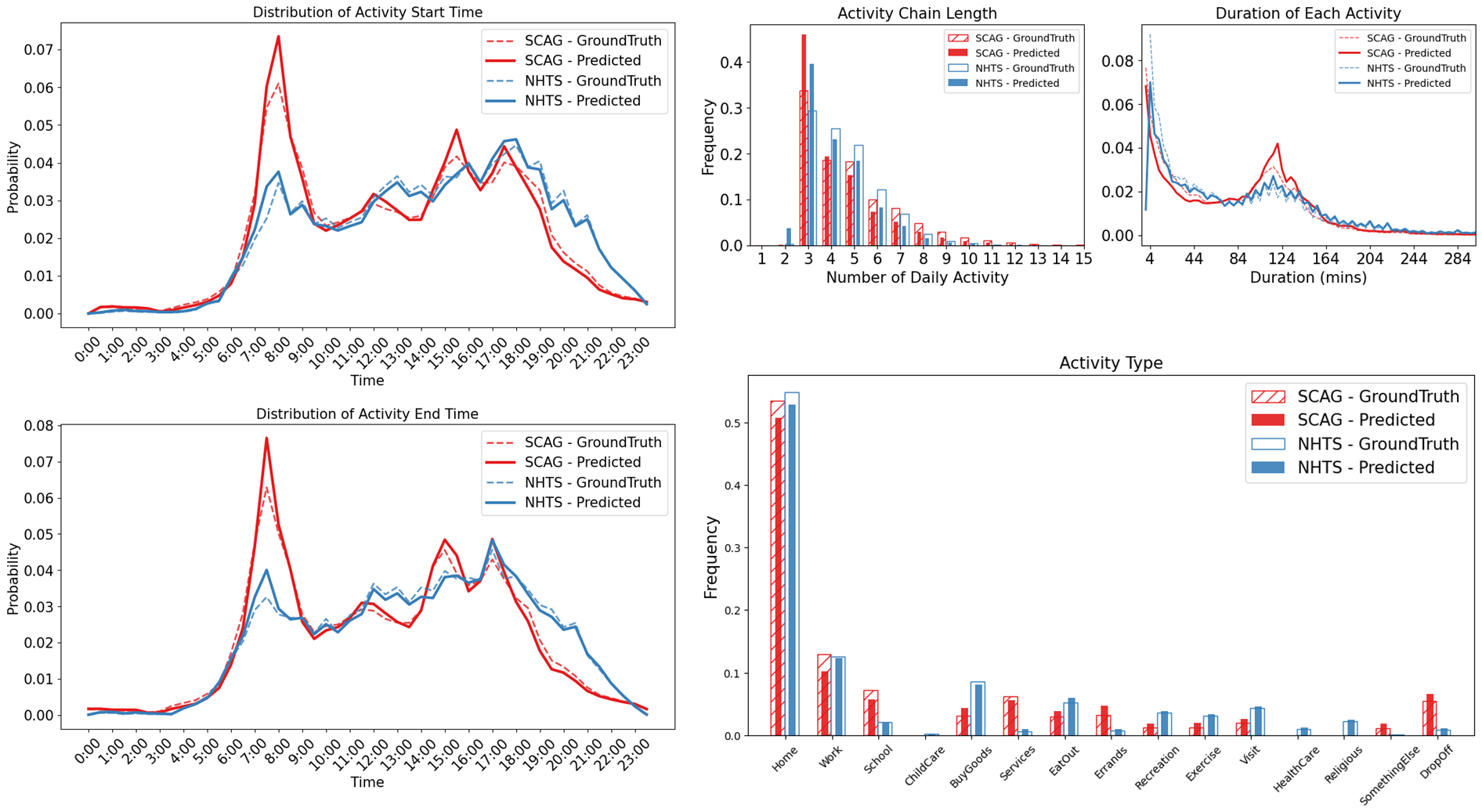}
  \caption{Performance evaluation on SCAG and NHTS dataset}
  \label{fig:SCAGNHTS_overall}
\end{figure}

\subsubsection{Base Model Evaluation}
Our initial evaluation is performed on the SCAG dataset and the NHTS dataset. The model's ability is tested by reconstructing 70\% masked activity chains from the test set, using three types of masking methods. The similarity of daily activity counts (chain length), activity duration, types, and start/end times between the reconstructed and original test set activities is measured. 

The base model $M^S$, trained on SCAG data, performs well when reconstructing masked SCAG data, as evidenced by the close alignment of red solid (predicted) and dashed (ground truth) lines in Figure \ref{fig:SCAGNHTS_overall}. This performance is quantified by JSD scores between 0.001 and 0.011 (Table \ref{table:jsd_scagnhts}, row 2). However, when applied to the NHTS dataset without modification, the model's performance declin significantly, with JSD scores increasing to 0.003-0.021 (Table \ref{table:jsd_scagnhts} row 3), This indicates that the model trained on SCAG data (Los Angeles area) doesn't adapt well to national patterns. 

\begin{table}[h]
\centering
\caption{JSD value for similarity measurement across datasets and model performance}
\label{table:jsd_scagnhts}
\begin{tabular}{|c|c|c|c|c|c|}
\hline
JSD & Length & Duration& Type & Start Time & End Time \\ \hline
SCAG vs NHTS & 0.017 & 0.009 & 0.059 & 0.018 & 0.013 \\ \hline
\textbf{$M^{s}$ reconstructs SCAG} & 0.011 & 0.004 & 0.004 & 0.001 & 0.001 \\ \hline
$M^{s}$ reconstructs NHTS & 0.021 & 0.003 & 0.017 & 0.004 & 0.007 \\ \hline
\textbf{$M^{n}$ reconstructs NHTS} & 0.017 & 0.002 & 0.001 & 0.001 & 0.005 \\ \hline
\end{tabular}

\end{table}

As shown in the first row of Table \ref{table:jsd_scagnhts}, there are significant differences between these two datasets. The JSD scores, ranging from 0.009 to 0.059 across various activity attributes, quantify these differences. For instance, as shown in Figure \ref{fig:SCAGNHTS_overall}, the activity start time distribution shows a notable peak for SCAG around 8:00 am, which is less pronounced in the NHTS data. The activity type graph reveals that SCAG data lacks representation in categories like childcare, health care visit, and religious activities, which are present in the NHTS dataset.

To address these discrepancies, we implement transfer learning for the model $M^S$, fine-tuning only the outer layers (embedding and MLP) of the model $M^S$ to capture nationwide patterns from NHTS data. This approach yields marked improvements, with the NHTS dataset fine-tuned model $M^N$ achieving JSD scores as low as 0.001 for activity types and 0.002 for durations (Table \ref{table:jsd_scagnhts}, row 4). Figure \ref{fig:SCAGNHTS_overall} visually confirms this improvement, showing close alignment between blue solid lines and blue dash lines lines across all distributions. These results demonstrate our algorithm's effectiveness in adapting to different geographical contexts and capturing diverse mobility patterns, despite the initial significant disparities between the SCAG and NHTS datasets.

\begin{figure}
  \centering
  \includegraphics[width=0.98\textwidth]{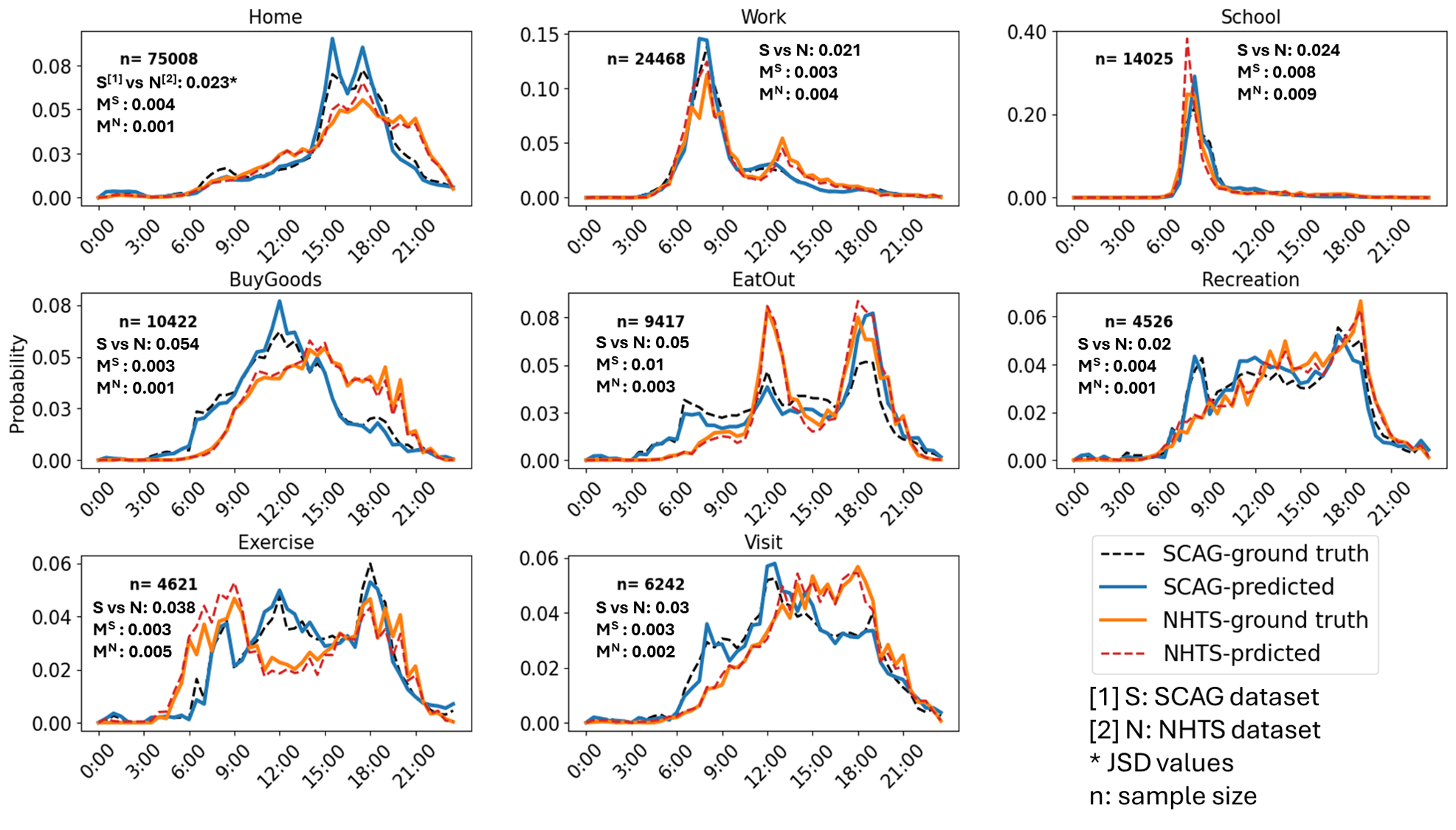}
  \caption{The start time of common activities reconstructed by base and its transferred model.}
  \label{fig:SCAGNHTS_activity}
\end{figure}

Beyond its commendable performance in system-level JSD evaluations, the proposed model exhibits proficiency in capturing patterns of common activities, with the transfer learning successfully adapting the model from regional to national contexts. As illustrated in Figure \ref{fig:SCAGNHTS_activity}, the model demonstrates remarkable accuracy in reconstructing the temporal patterns of activity start times across both datasets. 

Essential activities such as home, work, and school show closely aligned patterns between the two datasets, evidenced by low JSD values (0.023, 0.021, and 0.024 respectively), highlighting the universality of these routines. Conversely, regional variations are apparent in other activities. Shopping patterns, for example, differ significantly, with SCAG data showing a midday peak (JSD 0.054) versus NHTS's more even distribution, suggesting distinct regional shopping behaviors. Exercise activities also vary, with NHTS indicating a preference for early morning sessions (JSD 0.036). These differences likely mirror varying lifestyle choices between urban Los Angeles and the broader national context. The model's consistent low $M^S$ and $M^N$ values, typically below 0.01, confirm its efficacy in capturing these nuances, underscoring its utility for urban planning and policy-making by identifying both universal and regional behavioral patterns.

\subsection{Semi-Supervised Iterative Transfer Learning}
The domain adaptation capability of the propose training approach is validated using a sparse GPS trajectory data collected by Veraset \cite{veraset} from Egypt. The trajectory data is further processed using the trajectory annotation method shown in Figure \ref{fig:SystemWorkFlow}, converting trajectories to activity chains. To ensure data quality and representativeness, we implement stringent filtering criteria: only agents with a minimum of 7 days of recorded activity were retained, and we exclusively utilize samples that captured at least 25\% of daily temporal coverage. This rigorous selection process yields a substantial dataset of 479,526 samples, which we partition into training (80\%) and validation (20\%) sets. Notably, given the absence of ground truth data in this context, A separate test set is not allocated. The foundation for our transfer learning approach to the Egyptian Veraset data is the model trained on NHTS dataset, selected for its comprehensive coverage of 15 distinct activity types, thus providing a robust baseline for adaptation to the new geographical and cultural context, i.e., from the U.S. to Egypt.

\begin{figure}
  \centering
  \includegraphics[width=0.98\textwidth]{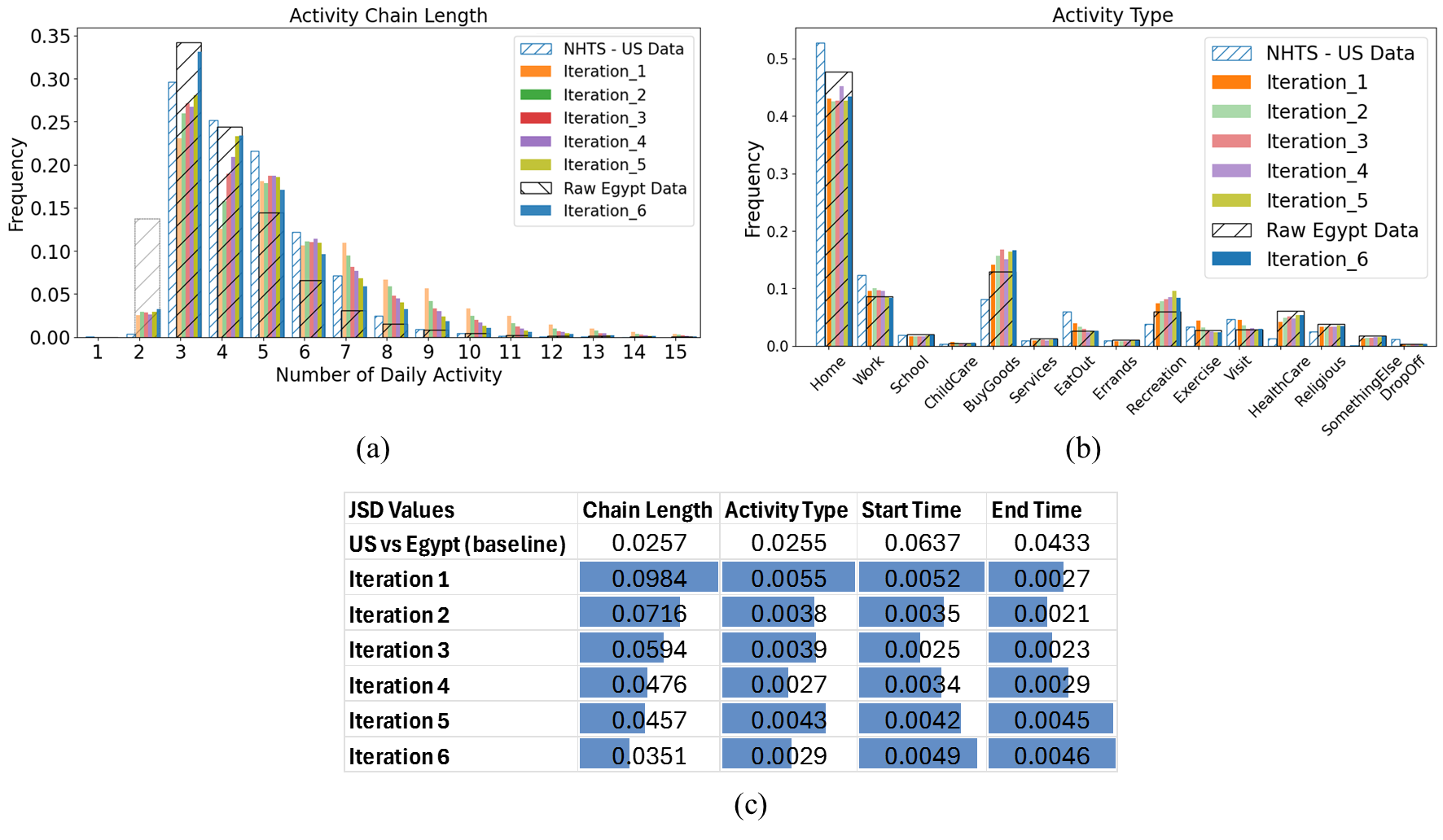}
  \caption{Performance evaluation: model evolution during iterative transfer learning. (a) Activity chain length, (b) type distribution, and (c) JSD value compared to raw Egypt data.}
  \label{fig:Transfer_TypeLength}
\end{figure}

\subsubsection{System-Level Analysis}
\begin{figure}
  \centering
  \includegraphics[width=0.98\textwidth]{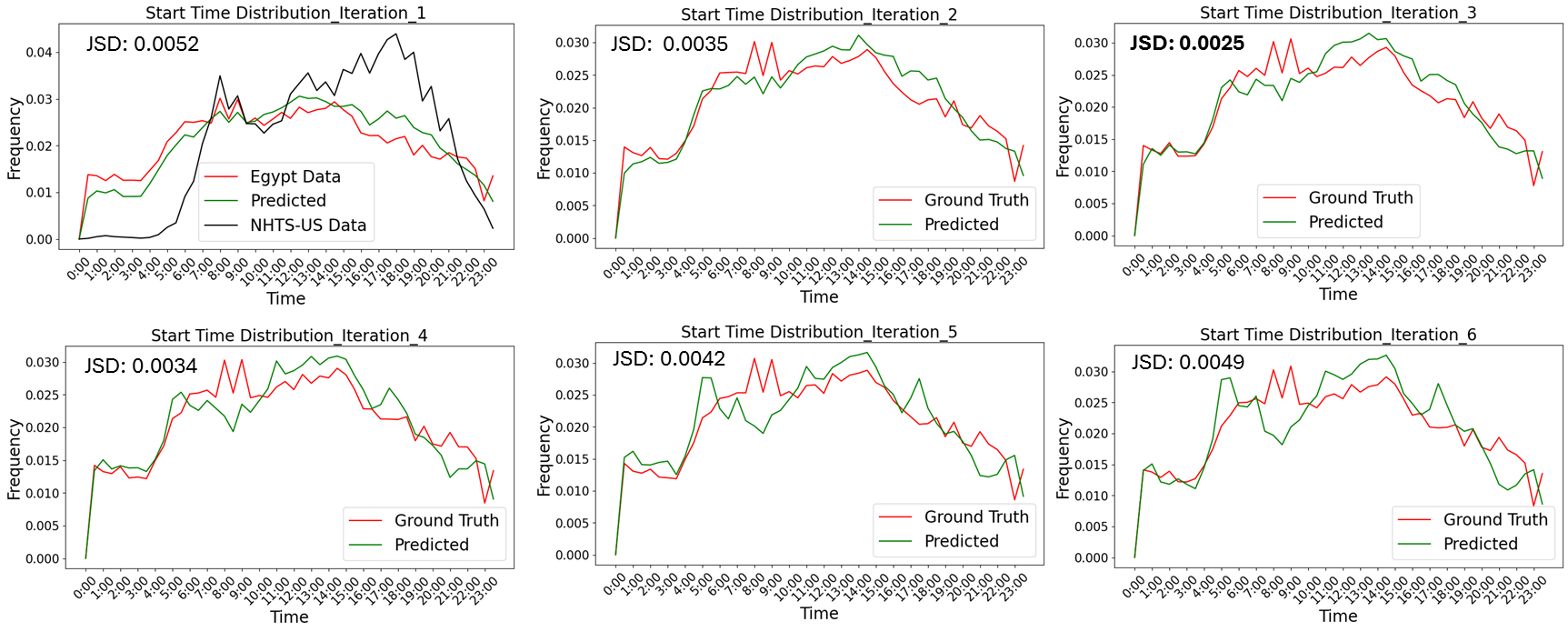}
  \caption{Model evolution during iterative transfer learning. Performance evaluation on start time distribution.}
  \label{fig:Transfer_ST}
\end{figure}

The Semi-Supervised Transfer Learning approach demonstrates a progressive adaptation of the NHTS-based model to the Egyptian context over multiple iterations, as illustrated in Figure \ref{fig:Transfer_TypeLength} and Figure \ref{fig:Transfer_ST}, which visualize the the adaptation process and provide quantitative results. 

The progression of the proposed approach in adapting the NHTS-based model to the Egyptian context over multiple iterations is elaborated in Figure \ref{fig:Transfer_TypeLength}. Figure \ref{fig:Transfer_TypeLength}(a) depicts the evolution of activity chain length distribution. Notably, the raw Egyptian data's high frequency of two-activity chains (blurred bar) indicates data incompleteness rather than true behavior. As iterations progress, the model shifts from the NHTS pattern towards a more realistic Egyptian distribution, peaking at 3-4 activities per day. This demonstrates the model's ability to infer plausible patterns from incomplete data. Figure \ref{fig:Transfer_TypeLength}(b) depicts activity type frequencies. Home and Work activities remain dominant across all iterations, reflecting their universal importance. The model successfully adjusts for lower frequencies of activities like buy meals, religious activity, and health care visit in the Egyptian context, while increasing the frequency of buy goods activities to match local patterns. As the iterations progress, a gradual shift is observed from the initial NHTS-US pattern towards a distribution more closely aligned with the raw Egyptian data, indicating successful adaptation to local activity patterns. 

The quantitative results are presented in Figure \ref{fig:Transfer_TypeLength}(c), which displays JSD values for the evaluation metrics across iterations. The first row of the table shows the dissimilarity between the U.S. (NHTS) dataset and the Egypt (Veraset) dataset, providing a baseline for comparison. Subsequent rows demonstrate the progressive adaptation through iterations. The "Length" column reveals a general decrease in JSD values from Iteration 1 (0.0984) to Iteration 6 (0.0351), indicating increasing similarity to the target distribution of activity chain lengths by \%64. Activity Type JSD improves rapidly, reaching a low of 0.0027 in Iteration 4. Start Time reaches its lowest at Iteration 3 (0.0025), while End Time is lowest at Iteration 2 (0.0021), however, these two JSDs increase in later iterations.

Building upon the previous analysis of JSD values, we further examine the temporal pattern (e.g., start time distributions) across iterations to gain deeper insights into the adaptation process. 
Figure \ref{fig:Transfer_ST} presents a series of graphs illustrating the start time distributions for each iteration. The initial iteration compares NHTS-US data, Egyptian data, and the model's predictions, while subsequent iterations focus on refining predictions against Egyptian ground truth. In Iteration 1, we observe a stark contrast between the NHTS-US data, which shows a pronounced mid-day peak, and the more uniformly distributed Egyptian ground truth with multiple smaller peaks. As iterations progress, the predicted distribution increasingly aligns with the Egyptian data, reaching optimal alignment in Iteration 3 (JSD: 0.0025). This iteration effectively captures the characteristic multiple peaks of Egyptian activity patterns, particularly during morning and evening hours. However, in Iterations 4-6, a gradual divergence occurs as the model overemphasizes certain peaks while underestimating others, resulting in a slight increase in JSD values from 0.0034 to 0.0049.

\subsubsection{Activities-Level Analysis}
\begin{figure}
  \centering
  \includegraphics[width=0.98\textwidth]{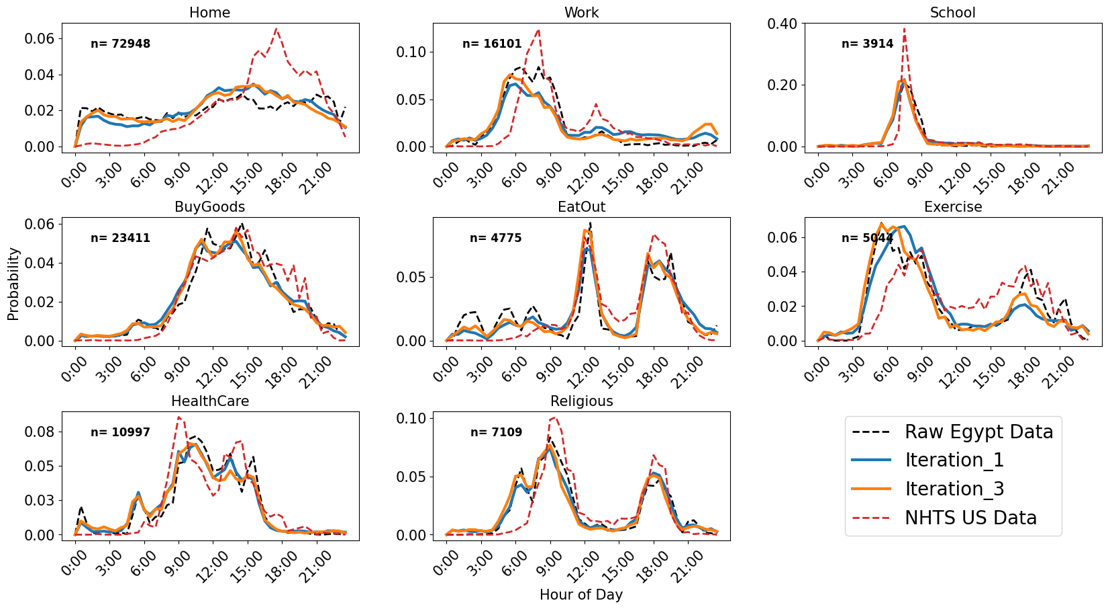}
  \caption{The comparison for start times of common activities in U.S., Egypt, and the model adaptation process}
  \label{fig:TransferActivityLevel}
\end{figure}

A detailed activity-level analysis is presented in Figure \ref{fig:TransferActivityLevel} and Table \ref{table:JSD_activities}, which compare the activity start time distributions across different activity types for raw Egyptian data, NHTS US data, and two iterations of the model's predictions. This activity-level analysis reveals the cross-cultural differences in daily activity patterns and the efficacy of the proposed iterative transfer learning approach in adapting to these differences.

\begin{table}[h!]
\centering
\caption{JSD values for start times of common activities  reconstruction in Egypt} \label{table:JSD_activities}
\footnotesize
\setlength{\tabcolsep}{5pt} 
\begin{tabular}{lcccccccc}
\toprule
 JSD values & Home & Work & School & Buy Goods & Buy Meals & Exercise & Health Care & Religious \\
\midrule
US vs Egypt (baseline) & 0.0954 & 0.0886 & 0.1068 & 0.0268 & 0.0999 & 0.0581 & 0.0546 & 0.059 \\
Iteration 1 & 0.0087 & 0.0449 & 0.0262 & 0.0055 & 0.0269 & 0.0218 & 0.0115 & 0.0134 \\
Iteration 3 & 0.0094 & 0.0372 & 0.0157 & 0.0051 & 0.0197 & 0.0169 & 0.0089 & 0.0195 \\
\bottomrule
\end{tabular}
\end{table}

\textbf{Mandatory Activities}: Home activities in Egypt show a more distributed daily pattern compared to the US's pronounced evening peak. As shown in Table \ref{table:JSD_activities}, this difference is quantified by a high baseline JSD of 0.0954, which is improved to 0.0087 by Iteration 1 and at 0.0094 in Iteration 3. Work activities start times exhibit a sharp morning peak in both datasets, but the Egyptian data shows a broader distribution, suggesting more varied work schedules, and the model progressively learns the pattern with JSD decreasing from 0.0886 to 0.0372 by Iteration 3. The peak time of School activity are similar across both datasets, and the model quickly capturing this similarity with JSD improving from 0.1068 to 0.0157 by Iteration 3.

\textbf{Non-Mandatory Activities}: The patterns of buy goods activities are similar between countries but with subtle timing differences, reflected in the model's quick adaptation and maintain stable through iterations (JSD: 0.0055 by Iteration 1 to 0.0051 by Iteration 3). This similarity is also presented in buy meal (eat out) activity. On the other hand, religious activities show stronger morning and evening peaks in Egypt, likely reflecting Islamic prayer times. The model successfully adapts to these culturally specific patterns, with the JSD improving from 0.059 to 0.0134 in Iteration 1, though slightly increasing to 0.0195 in Iteration 3, still showing good overall adaptation.

Overall, the model effectively adapts to Egyptian activity patterns, showing rapid progress from baseline to Iteration 1 and further refinement by two more iterations. This capability is evident both visually and through decreasing JSD values, indicating a close alignment with local data while maintaining influences from the NHTS data. This analysis highlights the model’s potential to generate realistic synthetic mobility data, reflecting local behaviors and cultural nuances, which is crucial for urban planning and cross-cultural behavioral studies.

\subsection{Limitation: Synthetic Data and Model Collapse}

Our transfer learning approach shows promise in adapting mobility patterns to the Egyptian context, but it also reveals limitations in later iterations. The deterioration in model performance beyond Iterations 2 and 3 aligns with recent findings by Shumailov et al.\cite{shumailov2024ai}, who reported that "AI models collapse when trained on recursively generated data". This phenomenon occurs when model-generated content is indiscriminately used in training, leading to the loss of nuanced or rare activities in the original data distribution.

The initial iterations successfully capture broad Egyptian mobility patterns. However, continued training on synthetic data risks amplifying minor inaccuracies and biases. This underscores that synthetic data, while valuable, is not a panacea for transfer learning. Early stopping, in our case at Iteration 3, appears to provide the optimal balance between local adaptation and data integrity.

This limitation highlights the ongoing importance of collecting data from genuine human interactions. Real-world data provides essential grounding that synthetic data alone cannot replicate. In conclusion, while our approach demonstrates potential in cross-cultural mobility modeling, it also reveals the complexities of relying heavily on synthetic data. Balancing data augmentation through transfer learning with the need for authentic, diverse data remains a critical challenge in mobility modeling and AI research.

\section{Conclusion and Future Work}
This study presents a novel model for reconstructing human mobility patterns by focusing on semantic activity chains. Our semi-supervised approach and transfer learning techniques enable the model to adapt to different regions and data availability scenarios, addressing the limitations of traditional supervised learning methods. The model's effectiveness is demonstrated through extensive validation with datasets from the United States and Egypt, showcasing its ability to generate realistic and high-quality synthetic mobility data. These findings highlight the model's potential as a powerful tool for urban planning, policy development, and transportation system analysis.

Future work will focus on enhancing the model by incorporating socio-demographic factors of agents to reconstruct more realistic activity chains for different types of people. This improvement aims to provide a more personalized understanding of mobility patterns. Additionally, the current model is limited to reconstructing semantic activity chains; therefore, future research will extend the model to include location reconstruction. This advancement will enable a comprehensive modeling of both the activities and their spatial contexts, further enriching the insights into human mobility patterns. These developments will enhance the model's utility for urban planning, policy development, and transportation system analysis across diverse geographical contexts and data-scarce environments.

\section{Acknowledgements}
Supported by the Intelligence Advanced Research Projects Activity (IARPA) via Department of Interior/Interior Business Center (DOI/IBC) contract number 140D0423C0033. The U.S. Government is authorized to reproduce and distribute reprints for Governmental purposes notwithstanding any copyright annotation thereon. Disclaimer: The views and conclusions contained herein are those of the authors and should not be interpreted as necessarily representing the official policies or endorsements, either expressed or implied, of IARPA, DOI/IBC, or the U.S. Government.

\newpage

\bibliographystyle{trb}
\bibliography{Reference}
\end{document}